# Existence and perception as the basis of AGI

**Victor Senkevich**                                                                                                     VICTORS@USA.COM
*Independent researcher & developer*
*L5L 3M2*
*Mississauga*
*Ontario*
*Canada*

## Abstract

As is known, AGI (Artificial General Intelligence), unlike AI, should operate with meanings. And that's what distinguishes it from AI. Any successful AI implementations (playing chess, unmanned driving, face recognition etc.) do not operate with the meanings of the processed objects in any way and do not recognize the meaning. And they don't need to. But for AGI, which emulates human thinking, this ability is crucial. Numerous attempts to define the concept of "meaning" have one very significant drawback — all such definitions are not strict and formalized, so they cannot be programmed. The meaning search procedure should use a formalized description of its existence and possible forms of its perception. For the practical implementation of AGI, it is necessary to develop such "ready-to-code" descriptions in the context of their use for processing the related cognitive concepts of "meaning" and "knowledge".
An attempt to formalize the definition of such concepts is made in this article.

## 1. Basic definition

**Definition:** Existence is the belonging of an element to a set.

This is a universal basic definition. Next, we develop this definition and draw some conclusions.

**Recursiveness:** the strength of this definition lies in its recursiveness. The existence of a set is also its belonging as an element to another set. i.e., this definition is applied recursively and allows us to understand and describe the complexity and diversity of various forms of existence.

**"What" and "Where":** all attempts to fix the existence of something will inevitably lead to the need to determine the 2 entities "what exists" — the element and "where it exists" (or "when", and for immaterial entities even "how") — the area of existence, the set to which this element belongs.

In addition to the values "exists" and "does not exist", the criterion of existence has an initial value of "undefined". This value unambiguously indicates the subjectivity of the concept of existence — "undefined" actually means "unknown" for some subject, an observer seeking to infer the existence of some object. for another subject, the value of the criterion of existence for a given object can be completely determined.

The category of existence is fundamental and inseparable from the category of truthfulness (nevertheless, these categories are subjective). Because only what exists is true. And the whole functioning of intelligence, whether natural or artificial, boils down to the problem of finding the truth in its various forms — answering questions, formulating statements, searching for solutions. As is known, "correctly asked questions already contain answers." A correctly asked question, first of all, defines and limits the area of existence of the answer. That is, a correctly formulated question reduces uncertainty. And this determines its "correctness".

In our formulation, the search for a solution /answer is the process of determining a set that is such an area of the existence of possible solutions / answers and then determining in it the fact of the existence of a specific answer to a given question.

## 2. Physical interpretation

It is quite understandable on an intuitive level that for material objects something can exist only in a certain area of space that encompasses it. Being inside means it exists for this area. And it does not exist for another one if these areas do not interact in any way (i.e. they themselves are or are not elements of another set in the same way — the definition is applied recursively). If 2 areas of space (defined as sets) do not belong to any other area, then the objects of one area (elements) do not exist for the other. This is an interpretation of this definition for material objects. In the universal interpretation, it is "set" and "element" that are used. Since the set is the most general entity that can include its constituent parts, elements. Space, territory, region, and even the Universe are less common such entities. Because even the Universe can only be an element in the set of all Universes. And since the objects themselves and the area of their existence can be not only a physical area of space, but also immaterial, virtual entities — these can be, for example, the parameters of the functioning of some system and the area of their possible values.

## 3. The existence of material and intangible

The universality of the proposed definition makes it possible to remove the contradictions that arise when trying to distinguish between the existence of material and intangible objects. The object is material or intangible, as well as the set to which it belongs. But for a given set, it exists — regardless of whether it is material. This approach allows us to consider constructively and uniformly existence of any material or intangible, abstract objects.

Intangible objects surround us everywhere — these are, for example, parameters of economic or social systems, abstract terms of mathematical theories, and even many seemingly simple concepts of spoken language. By asking the question "does this make sense?" we are trying to determine the existence of an immaterial object "sense" on a set of correct lines of behavior, to understand whether the proposed action belongs to a set of correct decisions, "meanings". Below we will define this term more strictly.

Thus, this basic definition links together subjectivist (relativistic, idealistic..) and objectivist ("materialistic") approaches to the problem of existence. For quite material sets, existence is material, for "intangible" existence is immaterial, but this is also existence. It is obvious that subjectivity plays an important role in understanding the concept of existence. The role of the subject (the observer of sets and the elements existing in them) is decisive, in fact, the observer determines (for himself, this is important) both the set and the tested element. Accordingly, it must be recognized that existence is a subjective and relative category.

## 4. Relations

**Definitiōn:** A relation is an interaction or connection of any kind between the elements of a set.

If 2 elements belong to a set, they may or may not be related by some relation. If they are not connected by any relation, they exist, but are not perceived by each other. Not observable. Not tangible. Not measurable. If the relation exists, i.e. the elements are connected by some relation, then such elements will be called perceived by each other (see the definition of perception below). Perception can be unidirectional, one-sided or bidirectional, two-sided. The existence of relations on a set is well illustrated by a disjoint directed graph.

The physical meaning of the relations is determined by the nature of the set. A man sees an apple. It (element) existed in the garden (set) before the man (element) saw it. After a man saw an apple, this element became perceived. A relation appeared on a set of elements. If a subject understands or perceives a certain concept as an element of a set, it exists to the same extent that the set itself exists as an element of some other set.

Any element of a set exists to the same extent as the set itself exists.

The square root of the number 2 exists on the set of real numbers if such a set itself exists. The beauty of a work of art exists only for the one who perceives it — feels the relation of preference on a set of such works. The tractor (element) in the field (set) exists and is perceived for the tractor driver (element), exists and is not perceived for the rest of the workers (elements) of the village (set) until they see it, and for everyone else it exists or does not exist, as well as the whole village as an element of their system of the universe (set).

## 5. Paradoxes are possible

Due to the set-theoretic nature and recursiveness of the proposed basic definition of existence, it admits paradoxes. It is also interesting to note that for some elements, the fact of existence can be determined by creating a relationship that removes (destroys, changes) an element from the set. That is, it is possible to determine that an element exists (to measure it) only by removing (destroying, changing) it. This is known to be observed in quantum physics. In the macrocosm, any observation (the implementation of a relationship) also changes both the observed object and the observer, it's just that such changes may be recognized as insignificant, but they always exist.

## 6. «Cōgitō ergō sum»

As it was shown above, the presence of relations between two elements of the set makes them perceived by each other. Accordingly, it fixes the fact of the existence of one object of perception for another. If such a relation is unidirectional, the fixation of existence occurs only for one object. A photo of an artist in a soap advertisement captures the fact of his existence (both artist and soap) for many people, but for the artist (and soap) the fact of the existence of specific people who have seen the advertisement remains unknown. The relation can also be applied to a single element. In this case, the object's existence is fixed only for itself.

"Cōgit er ergō sum" — "I think, therefore I am" (Descartes, R. 1637), the statement of Rene Descartes illustrates this fact. At the same time, taking into account the above, it can be concluded that Descartes' statement is correct, since any statement of an object relative to itself is already a relation and, accordingly, a confirmation of existence (however, only for itself). But it should be noted that this statement is partial, not general, applicable only to a part of existing entities, namely, thinking beings. Since "I do not think, therefore I exist" and many other statements will also be true (for example, "I Am Groot". Groot only knew this phrase and pronounced it, proving his existence as a thinking person to himself and others). The brief statement "I am" is already a relation and, accordingly, a perception of oneself and a confirmation of existence.

## 7. Perception, data, meaning and knowledge

**Definition:** Perception is the projection (result of interaction, realization of connection) of a relation onto an object (element(s) of a set).
Consciousness is perception. Self-Consciousness is the perception of one's own essence.
Attention is a form of perception. As well as touch, smell, hearing. And even consideration can be called a form of perception of the object of consideration.

**Definition:** A datum is a representation of any kind (for example, a description or a machine view) of a single element of perception.
Thus, a datum is a representation in any form of the elementary result of the interaction of some objects.

**Definition:** Any selection of datums forms data (dataset).

**Definition:** Meaning is a representation of any kind (for example, awareness or description, including formula, algorithm, program code) of a single act of relation.
Thus, meaning is the representation in any form of a single act of interaction of some objects as a whole, and not just the result of such interaction. I.e., meaning in our terminology is the mechanism of a single interaction, and data is the result of such interaction. The uniqueness of interaction can be understood as the integrity, completeness of the elementary act of interaction. Due to the recursiveness of everything considered in this article, a "single" interaction can also be a rather complex result of previous recursions.
Based on the above definitions, the question considered above "does it make sense?" can be presented as an attempt to determine whether the intended action (a single act of a relation) belongs to a holistic view of the interaction of objects (some relation). An action may make sense, i.e. correspond to some interaction, for example, the collaboration of employees to achieve a group goal, or it may not make sense if the action does not belong to a set of correct actions.

**Definition:** Intelligence is an operator of meanings. A subject operating with meanings, forming, creating meanings, i.e. determining (for oneself, since meanings are subjective) the existence of relations between elements of various sets, environmental objects or virtual entities.

**Definition:** Knowledge is a certain set of meanings as a representation of any kind (for example, awareness or description, including formula, algorithm, program code) of a relationship as a whole or any subset thereof.

**Definition:** Truth is what exists.
And truthfulness is existence. Existent is true. Everything that is true exists. Knowledge is existing. Knowledge is true. The formation of meaning by the intellect is the determination of truthfulness, i.e. the fact of existence.

Thus, knowledge is a set of representations (meanings) about relations - interactions or connections between objects of any nature. And data is a representation of the perception of the results of such interaction.

**Example:** for the number π:

- the number π belongs to the set of real numbers and, in this sense, exists. To the same extent as the set of real numbers itself
- the definition of the number π in any form is a relation, it defines the relation of the number π with other real numbers
- any formula for calculating the number π is knowledge
- the formula for calculating any digit of the number π is the meaning
- any calculated decimal place of the number π (or its approximation to this decimal place) is a datum
- any subset of the decimal places of the number π is data

**Remark:** Data is, of course, information. In this article we do not consider the connection of these concepts in the context of Information theory. But it is obvious that both data and knowledge are also associated with a level of uncertainty, or Shannon entropy. Our goal, among other things, is to show the subjectivity of this concept. In Information theory, the level of uncertainty is quite an objective value — in a strictly defined very abstract area. In our interpretation, the objectivity, materiality and very existence of any value directly depends on the area in which it is defined.

## 8. A System and Ontology as a Function of Perception

**Definition:** A system is an object with diversity.
The "object" means that the system can somehow be isolated from the surrounding world through perception. "Diversity" means that the system has a perceived structure — attributes, behaviors, elements, etc.

The perception of diversity in the surrounding world is the environment for the formation of ontologies.

**Definition:** Ontology is metadata.

The prerogative of the intellect, whether natural or artificial, is to create its own ontologies, interacting with the surrounding world (subject area) and determining its structure, individual for a given subject. The structure of the subject area is described by metadata representing knowledge (in the simple case, metadata is data about data, also representing a relation, i.e.

knowledge). Where such a mechanism is present — there is intelligence. Where it is rigidly set from the outside, there is a software robot.

In this sense, any material object — a stone, a table, an apple — becomes a system when we notice the presence of diversity of any kind in them: elements of ontology, attributes, categories, properties (colors, smells, other physical parameters). It's the same with intangible objects. The text of Leo Tolstoy's novel "War and Peace" and Shostakovich's symphony are not at all diverse, but, on the contrary, quite monotonous for an observer who does not have the perception of such diversity. The random number generator does not have variety and complexity. And it is not a system. But for someone who found out that some such generator was implemented using the function of calculating the number π, it already is. Both the text of a literary work and a symphony can be perceived as such a generator — a random sequence of letters or notes, fairly evenly distributed. These works become a system only for the observer who perceives their structure (including meaning, harmony). For the rest of the people, this is not a system, but information noise or sound noise.

Thus, the qualification of an object as a system depends on the perception of its inherent diversity (some call it complexity, structure, properties, connections, but diversity is the most general term). It does not matter whether the object in question is static or dynamic. A dynamic entity is not a system until you imagine it as such and realize that it has a structure, connections, behavior. And until you begin to perceive it as a system. An apple that has fallen from a tree onto someone's head is not a system for most, it's just a dynamic object. And for Newton, who imagined a set of connections and properties accompanying an apple, it is such.

## 9. Triggers, Motivations and Free Will

Perception can initiate triggers. Attention, touch, sensation, and even thinking as forms of perception are tools for the formation of triggers.

**Definition:** Triggers are reactions of the subject.

**Definition:** Motivation is a conscious need for action.

Triggers are simply reactions of the subject, but we are in favor of using more formalized terms for AGI instead of biological ones. Triggers form desires, goals and the need for action, motivation. A question asked, a ball thrown, a feeling of hunger or thirst, a thought that arose in the process of thinking, etc. are perceived using the available tools (forms of perception). Or they are not perceived and then the triggers do not arise. The thrown ball flies past. The interlocutor's thought is not perceived. If perception (attention, feeling, etc.) detects a deviation/event, then a trigger (reaction) is fires that forms a goal or desire, motivation (dodge or catch a ball, drink water, answer a question, etc.) Further, when generating a response, uncertainty arises or does not arise (how to catch the ball, where to drink water, etc.).

**Definition:** Free will is the freedom of the algorithm to choose a solution in the presence of alternatives, using either built-in motivation or a random selection method in the absence of it.

## 10. Tasks, Goals and Dreams

In the general case, the Goal is a Task that has uncertainties that do not allow it to be unambiguously solved. Uncertainty is a fundamental property of the Goal, which makes it impossible to implement the Goals in the information system using any rigid algorithms. The goal, devoid of uncertainty, turns into a Task. And the Goal, devoid of certainty, turns into a Dream.

**Definition:** If the actions to achieve a certain state are completely clear and provided with resources, then this is a Task.

**Definition:** If the actions are completely clear and not provided with resources, then this is the Goal.

**Definition:** If the actions are not completely clear and not provided with resources, then this is a Dream.

**Example:** buying a car:

• If you have money, a driver's license, the availability of the right model in the store — this is a Task, it can be completed.

• If there is no money or no driver's license or no desired car model in the store — this is a Goal, it can be achieved in different ways by setting Tasks (getting a license, taking out a loan, etc.)

• If there is nothing but only desire, this is a Dream, we can dream about it or set intermediate Goals.

The presence of uncertainty makes Goals and Problems similar. The difference is that usually the Goals are formulated by the subjects themselves, and Problems, as a rule, arise from the outside.

## 11. "Yes", "No", "I don't know"

When creating a real AGI, it is necessary to determine the tools for constructing the derivability of statements in natural language. Applying the apparatus of formal logic for this, the first thing that must be immediately abandoned is the law of the excluded middle — the basis of binary logic.

A fundamental requirement is the need to handle uncertainties. Therefore, it is necessary to use three — valued logic (3VL) with a tuple of base values **{"yes","no","unknown/indeterminate"} / {true, false, null}.** Otherwise, artificial intelligence will never be able to process correctly questions like "Have you stopped drinking cognac in the morning? Yes or no?". This is the fundamental problem of the deductibility of propositions in natural language.

Three-valued logic allows to build iterative and recursive procedures for forming propositions in a more natural way. The values "**yes**" or "**no**" complete the process of cognition. The value "**unknown/indeterminate**" initiates a new stage of the process.

**Remark:** This requirement is basic. The toolkit can be extended using fuzzy and/or multi-valued logic, but the principle of not making a decision instead of making a wrong decision can be taken as an axiom when building AGI.

## 12. Existence, abduction and the search for meaning

"Sherlock Holmes practised abduction, not deduction" (Carson, D. 2009).

"Induction never can originate any idea whatever. No more can deduction. All the ideas of science come to it by the way of Abduction…Deduction proves that something must be. Induction shows that something actually is operative. Abduction merely suggests that something may be" (Peirce, 1934).

Abduction is a procedure for making an assumption that an element belongs to a certain set by the criterion of whether the element in question and all elements of the set have a common property. Abduction is a generalization procedure, creating a representation of a set based on known data about its elements. The very fact of the presence of a common property is not a proof of the element belonging to the set, as well as the existence of the set itself. But it allows us to make such an assumption. Iterative application of such a procedure is a way (not guaranteed) to verify the validity of the assumption being tested. The common property of the elements of a set is nothing more than a representation of a relation in our more general definition. That is, the meaning in its definition given above. Thus, abduction can be used as a tool for searching for meanings, i.e. elements of knowledge. Which, in fact, was claimed by the author of the term abduction (Pierce, 1934), who believed abduction to be a mechanism for the emergence of scientific hypotheses. We just use our set-theoretic approach to the concepts of existence and perception to show that consecutive application of abduction, the meaning search procedure can be formalized. And, possibly, used in the development of Artificial General Intelligence (AGI).

## 13. Aristotle's syllogisms as a logical processor for AGI

Abduction can be used in AGI for the most important procedure of generalization and formation of Abduction can be used in AGI for the most important procedure of generalization and formation of meaning. Deduction and induction are no less important for the procedures of forming a logical conclusion, to operate with logical statements. In this article, we consider AGI in its less broad interpretation – as an artificial intelligence operating in a natural language.

For the first time the systematization of logical statements was carried out by Aristotle. Aristotle's syllogisms are a set of verbal statements that formulate the provisions of binary logic in natural language. Despite the fact that the apparatus of binary (boolean) logic is quite developed, including the form of set-theoretic operations, in this context we are interested in the implementation of logical statements in natural language. Therefore, Aristotle's syllogisms are very promising for use as an AGI logic processor, i.e. the apparatus for forming reasoning, logical conclusions and fixation of cause-and-effect relationships in natural language texts.

## 14. Patterns of meaning

In the natural language, which AGI should be able to operate with, relations (meanings and knowledge in the understanding and definition given above) are expressed by sentences containing logical statements, the truth of which is verified (proved) by sentences containing information, the truth of which does not need to be proved (data in the understanding and definition given above).

It is well known that there are a very limited number of types of possible syllogisms, statements in the form "**<subject> <predicate> <object>**" and possible variants of the structure of a declarative sentence. Thus, we can assert that the "AGI logical processor" must also operate with only a limited number of patterns of possible inferences (meanings).

This obvious conclusion is very important. It allows us to assert that even a very complex and intelligent AGI can be the result of recursion of a very simple basic model operating with simple logical statements, initially "charged" with some minimal set of such "patterns of meaning".

The addition of the "AGI logic processor" with the "abduction pattern" will allow it to build relations (knowledge) that generalize the information already available in its knowledge base. In practice, the task of the "basic" version of AGI can be formulated somewhat simpler — it is necessary not to construct new previously unknown relations (knowledge), but to test hypotheses asked by a person in the form of questions in a dialogue with AGI.

**Example:**
Relation (question asked by a person): "Have people been to the Moon?"
Data (information stored in the knowledge base): "American astronauts flew to the Moon"

This is the simplest example to demonstrate the above.
The question (hypothesis) and the available data are statements represented by a "pattern of meaning" in the form "<subject><predicate><object>". The "AGI logical processor" must make a conclusion about the truth or falsity of the statement being verified, i.e. the existence of a relation, knowledge in the above definition.

It is necessary to use abduction (which in this example can use semantics) to form the necessary generalizations ("astronauts are people", "they flew, so they were"). Once again, I note here that the use of the abductive method does not guarantee the unambiguity of the result — in practice, there is often a need for additional data to assess the reliability. Even in the simplest example under consideration, the abductive conclusion "they flew, so they were" is not absolutely reliable, because there is an option "They flew but did not reach".

So, "patterns of meaning" allow us to categorize the types of logical statements that AGI operates with and consistently plan the development of AGI versions, starting with the simplest and most amenable to algorithmization.

Also, "patterns of meaning" expand Boolean logic, since, in addition to strict logical statements like "Socrates is also a man", they cover any other statements containing uncertainties, ambiguities and contradictions. And dealing with uncertainties and ambiguities is the most important characteristic of intelligence.

## 15. Applicability

"Philosophers have only interpreted the world in various ways. The point, however, is to program it" – inspired by Marx and Engels (1888)

This is our own approach to solving the problem of existence, perception and their connection with cognitive concepts. We call it constructive, or formalized philosophy. This approach is characterized by the desire for rigor and brevity of definitions and adherence to the Occam principle.

**Remark:** Occam's razor principle is nothing more than a heuristic method of using abduction. The set-theoretical interpretation of cognitive concepts also allows us to formalize such an important philosophical concept as the hermeneutic circle — in our interpretation, this is nothing more than a recursive process of knowledge formation as a relations on a set of elements and the set itself based on the perception of data using abduction. We note in this regard that such an interpretation shows how closely the concepts of Occam's razor, abduction and the hermeneutic circle are related. All these concepts declare cognition as a process of sequential generalization, the formulation of general dependencies on the basis of partial, incomplete data.

**Remark:** The philosophical concept of "thing-in-itself", separated from the concept of "phenomenon", is nothing more than the formalized concept of "existence" defined in this work and the concept of "perception" separated from it. And various philosophical interpretations of the connection of these concepts are nothing more than informal attempts to describe the concept of "relation" defined here.

The formalization of philosophical concepts allows them to be programmed, quantified and applied to determine qualitative cognitive concepts, which is fundamentally important for the development of Artificial General Intelligence (AGI).

The application of this approach can be very different and not only in AGI. For example, as an alternative to inefficient statistical methods in some analytical and predictive tasks. A qualitative assessment of the state of economic or production systems can be tried to build not only on statistics of changes in system parameters, but on the definition of sets of parameters and the evaluation of the existence in such sets of parameters with a certain value. But in general, such a methodology fits well into the practice of using artificial intelligence (AI) systems and can be very promising for analytical and predictive AI implementations. The Artificial General Intelligence (AGI), which formulates logical statements in natural language, can be built on the basis of the principles outlined. It is well known that experts assessing the state of economic or production systems analyze quantitative parameters, but for their conclusions about the state of the system use qualitative assessments that have developed as a result of practical work and life experience. The algorithm for generating such qualitative estimates is very important for AGI and, it seems, can also be reproduced using the above approach.

## References

Psillos, S. 2004. "Handbook of the history of logic: Inductive logic. An explorer upon untrodden ground: Peirce on abduction."

Peirce, CS, Hartshorne, C and Weiss, P 1934, Collected papers of Charles Sanders Peirce: Vol. 5: Pragmatism and pragmaticism, Cambridge, MA, Harvard University Press.

Carson, D. 2009. "The Abduction of Sherlock Holmes". International Journal of Police Science & Management. Volume 11 Issue 2

Smith, Robin. 2020. "Aristotle's Logic". The Stanford Encyclopedia of Philosophy (Fall 2020 Edition)

Marx, K. and Engels, F. 1888. "Ludwig Feuerbach and the End of Classical German Philosophy". "Die Neue Zeit", issues No. 4 and 5.

Descartes, R. 1637. "Discourse on Method". Leiden, Netherlands